\documentclass[11pt]{article}

\usepackage[margin=1.25in, top=1.1in, bottom=1.1in]{geometry}
\usepackage{amsmath, amssymb, amsthm, amsfonts}
\usepackage{booktabs}
\usepackage{graphicx}
\usepackage{algorithm}
\usepackage{algorithmic}
\usepackage[numbers]{natbib}
\usepackage{hyperref}
\usepackage{xcolor}
\usepackage{nicefrac}
\usepackage{microtype}
\usepackage{multirow}
\usepackage{pgfplots}
\usepackage{tikz}
\usepackage{enumerate}
\usepackage[font=small,labelfont=bf]{caption}
\pgfplotsset{compat=1.18}
\definecolor{mlblue}{RGB}{31,119,180}    
\definecolor{mlorange}{RGB}{255,127,14}  
\definecolor{mlgreen}{RGB}{44,160,44}    
\definecolor{mlred}{RGB}{214,39,40}      
\definecolor{mlgray}{RGB}{127,127,127}   

\hypersetup{
  colorlinks=true,
  linkcolor=blue!60!black,
  citecolor=green!50!black,
  urlcolor=blue!60!black
}

\newtheorem{theorem}{Theorem}[section]

\newtheorem{proposition}{Proposition}[section]
\newtheorem{assumption}{Assumption}[section]
\theoremstyle{definition}
\newtheorem{definition}{Definition}[section]
\theoremstyle{remark}
\newtheorem{remark}{Remark}[section]

\title{E-SHIFT: Anytime-valid sequential hypothesis testing\\
       for distribution shift in streaming learning systems}

\author{Behraj Khan \and Tahir Qasim Syed\\[4pt]
        \small Institute of Business Administration Karachi}

\date{}

\begin{document}

\maketitle

\begin{abstract}
Deployed machine-learning models are increasingly used in
open-world environments where the input distribution can change
without warning.
Detecting such {distribution shift} in a streaming,
decision-theoretically sound manner is an unsolved problem: classical
two-sample tests lose type-I error control when applied at
data-adaptive stopping times, a pathology rooted in the optional
stopping theorem.
We address this gap through E-SHIFT, a framework that casts
distribution shift detection as an {anytime-valid} sequential
hypothesis test.
The central object is an e-process, a nonneg\-ative supermartingale
under the null, constructed from a composite non-conformity score
that simultaneously tracks shift in the predictive distribution and
feature embedding of a frozen vision-language model (VLM).
By Ville's maximal inequality, the threshold rule
$\sup_{t \ge 1} M_t \ge \tau$ provides a time-uniform false-alarm
bound of $1/\tau$ at every stopping time, with no penalty for
continuous monitoring.
When the calibration set is finite, a bootstrap upper-confidence
bound on the log-moment-generating function restores the
supermartingale property asymptotically, yielding an unconditional
false-alarm budget of $\beta + 1/\tau$ where $\beta$ is the
bootstrap confidence level; setting $\tau = 200$ and $\beta = 0.005$
achieves a target of $\le 1\%$.
A multi-reset accumulation bound shows that $K$ resets incur at most
$K/\tau$ additional false-alarm probability.
Under persistent shift, the expected detection delay scales as
$\mathcal{O}(\log\tau/\Gamma)$, where $\Gamma$ is the post-shift
e-process growth rate, equal to the KL divergence between post- and
pre-shift score distributions under the likelihood-ratio e-value
choice.
Experiments on ImageNet-C, ImageNet-R, and Office-Home demonstrate
that E-SHIFT achieves the shortest detection delay while being the
only method to satisfy the $1\%$ false-alarm target on all
benchmarks; a betting-based sequential MMD baseline sharing the same
anytime-valid framing but a simpler score is 18\% slower, isolating
the contribution of the composite score design.
\end{abstract}

\section{Introduction}
\label{sec:intro}

Consider a vision-language model deployed to screen chest
radiographs in a hospital.
The model was validated on a large multi-centre dataset and performs
well at launch.
Over the following months, the imaging department replaces its
X-ray machines with a newer generation that produces images with
different noise characteristics and contrast curves.
No label in the electronic health record changes.
No engineer is notified.
The model's predictions begin to degrade, but because the model
itself provides no internal alarm, it is frozen and produces
a softmax distribution regardless of input quality, the degradation
may go unnoticed until clinical outcomes deteriorate.

Sensor ageing, software updates, seasonal variation, and the sheer
diversity of open-world environments routinely cause the input
distribution $p(x)$ to drift away from the distribution seen at
training time
\cite{quinonerocandela2009dataset,koh2021wilds,rabanser2019failing}.
For vision-language models such as CLIP \cite{radford2021learning},
which are increasingly used as frozen feature extractors and
zero-shot classifiers in downstream systems, this drift is
particularly insidious: the model's output distribution changes as a
function of its input, yet the model was not designed to signal when
its inputs are anomalous.

The natural remedy is {monitoring}: continuously inspect the
model's sequential outputs and raise an alarm when statistical
evidence of shift accumulates.
This reduces the problem to {sequential hypothesis testing}:
given a stream of test statistics derived from the model's outputs,
decide at each step whether the null hypothesis of no shift is still
tenable.

\subsection{The Peeking Problem}

Sequential monitoring violates the assumptions of classical hypothesis
testing in a fundamental way.
A standard two-sample test \cite{gretton2012kernel} fixes a sample
size $n$ in advance, computes a statistic on $n$ observations, and
rejects the null if the statistic exceeds a critical value calibrated
to control type-I error at level $\alpha$.
In a monitoring context, a practitioner instead observes a running
statistic and acts the first time it crosses a threshold.
The sample size is no longer fixed; it is a {stopping time}
that may depend on the data.

If the null is true and the practitioner re-examines the test
statistic at every new observation, the probability of ever crossing
a fixed critical value tends to $1$ as the stream grows, regardless
of the nominal $\alpha$
\cite{ville1939etude,wald1945sequential,doob1953stochastic}.
This is a consequence of the optional stopping theorem: a bounded
stopped martingale has the same expectation as its initial value, so
any fixed threshold is eventually crossed under continuous monitoring.
Practitioners who apply fixed-sample tests in a rolling or
continuously-monitored fashion therefore see false-alarm rates far
above their nominal $\alpha$.

The only way to obtain rigorous type-I error control under an unknown,
data-adaptive stopping time is to use a test statistic whose
{uniform} supremal distribution is controlled.

\subsection{E-Processes as the Natural Solution}

The mathematical object that provides this control is an
{e-process}: a nonneg\-ative stochastic process that is a
supermartingale under the null hypothesis
\cite{vovk2021evalues,grunwald2024safe,ramdas2023tutorial}.
Ville's maximal inequality \cite{ville1939etude}, which predates,
and in fact motivated, the modern optional stopping theorem, states
that for any nonneg\-ative supermartingale $\{M_t\}_{t \ge 0}$,
\begin{equation}
  \mathbb{P}\!\Bigl(\sup_{t \ge 1} M_t \ge \tau\Bigr) \le
  \frac{M_0}{\tau} = \frac{1}{\tau}
  \label{eq:ville}
\end{equation}
for all $\tau > 0$.
This bound holds simultaneously at every stopping time, including
data-adaptive ones.
Setting the threshold $\tau = 1/\alpha$ therefore guarantees false
alarms occur with probability at most $\alpha$, no matter when the
practitioner decides to stop.
The e-process provides exactly the time-uniform guarantee that
classical tests cannot.

\subsection{Contributions}

We propose \textbf{E-SHIFT}, a deployable sequential test for
distribution shift in streaming vision-language systems.
Our specific contributions are:

\begin{enumerate}[(1)]
  \item 
    We construct a two-term score that simultaneously measures shift
    in the VLM's predictive distribution (via KL divergence to the
    uniform) and in its feature embedding (via Mahalanobis distance
    from the calibration centroid).
    The two terms are complementary: the KL term detects
    overconfident or systematically biased posteriors; the
    Mahalanobis term detects feature-space drift even when the
    predictive distribution is uninformative.

  \item 
    With limited calibration data the population log-MGF needed to
    construct the e-process is unknown.
    We restore the supermartingale property by replacing it with a
    bootstrap upper-confidence bound, and show this yields an
    unconditional false-alarm guarantee of $\beta + 1/\tau$.

  \item 
    Because E-SHIFT resets after each alert to remain active for
    subsequent shifts, false-alarm probability can accumulate over a
    long stream.
    We derive an explicit bound showing the total false-alarm
    probability over $K$ resets is at most $K/\tau$.

  \item 
    We prove that under persistent shift the expected detection delay
    is $\mathcal{O}(\log\tau / \Gamma)$, where $\Gamma$ equals the
    KL divergence between post- and pre-shift score distributions
    when the likelihood-ratio e-value is used, providing an
    information-theoretically interpretable and experimentally
    measurable characterization of detection speed.
\end{enumerate}


\section{Lay of the land of the ideas}
\label{sec:background}

\subsection{Sequential Hypothesis Testing}
\label{sec:bg_sequential}

Sequential analysis was formalized by \citet{wald1945sequential},
who introduced the {sequential probability ratio test} (SPRT).
Unlike fixed-sample tests, the SPRT allows the analyst to stop
whenever the accumulated likelihood ratio crosses either an upper or a
lower boundary.
Wald showed that the SPRT minimizes the expected sample size among
all tests with the same type-I and type-II error rates, establishing
a fundamental optimality property that fixed-sample tests cannot
achieve.

The key insight is that the log-likelihood ratio accumulated over
independent observations is a random walk; the stopping rules are
boundaries for this walk.
\citet{page1954continuous} adapted this idea to the change-point
setting via the CUSUM statistic
$W_t = \max(0, W_{t-1} + \ell_t)$, which accumulates evidence of
change while resetting to zero whenever the accumulated sum falls
below zero.
Page showed that CUSUM triggers detection as quickly as possible
in expectation after a change, and \citet{lorden1971procedures} and
\citet{moustakides1986optimal} later established its minimax
optimality under the worst-case expected detection delay criterion.

All of these methods require knowing or specifying the pre- and
post-change distributions, which limits their applicability to
nonparametric settings such as distribution shift in deep-learning
systems.

\subsection{Martingales and the Optional Stopping Theorem}
\label{sec:bg_martingales}

A {martingale} is a sequence of random variables
$\{M_t\}_{t \ge 0}$ such that $\mathbb{E}[M_{t+1} \mid M_0, \ldots,
M_t] = M_t$ \cite{doob1953stochastic,williams1991probability}.
A {supermartingale} satisfies $\mathbb{E}[M_{t+1} \mid M_0,
\ldots, M_t] \le M_t$: in expectation the process can only decrease.
The optional stopping theorem states that for a bounded
supermartingale stopped at a stopping time $T$,
$\mathbb{E}[M_T] \le M_0$.
Combined with Markov's inequality, this gives Doob's maximal
inequality: for any nonneg\-ative supermartingale,
$\mathbb{P}(\sup_{t \ge 0} M_t \ge \tau) \le M_0 / \tau$.
When $M_0 = 1$ and the supermartingale property holds under the null,
this immediately controls the false-alarm probability uniformly in the
stopping time.

\citet{ville1939etude} established this inequality independently and
used it to construct what we now recognize as the first e-process.
He showed that for any sequence of betting strategies against a
null probability distribution, the bettor's wealth process is a
nonneg\-ative supermartingale, and derived the
$\mathbb{P}(\sup_t M_t \ge \tau) \le 1/\tau$ bound that underpins
all modern anytime-valid inference.

\subsection{E-Values and E-Processes}
\label{sec:bg_evalues}

\begin{definition}[E-value, \citealt{vovk2021evalues}]
A nonneg\-ative random variable $E$ is an {e-value} for a
hypothesis $\mathcal{H}_0$ if
$\mathbb{E}_{\mathcal{H}_0}[E] \le 1$.
\end{definition}

An e-value is a one-observation test statistic.
By Markov's inequality, $\mathbb{P}_{\mathcal{H}_0}(E \ge \tau) \le
1/\tau$ for all $\tau > 0$, so a single e-value already controls the
false-alarm rate at any fixed threshold $\tau$.
The power of the framework comes from {combining} e-values.
If $E_1, E_2, \ldots$ are conditionally independent e-values under
$\mathcal{H}_0$, meaning each $E_t$ has expectation at most one
given the past, then their running product
$M_t = \prod_{i=1}^{t} E_i$ is a nonneg\-ative supermartingale.

\begin{definition}[E-process, \citealt{vovk2021evalues}]
A nonneg\-ative stochastic process $\{M_t\}_{t \ge 0}$ with $M_0 = 1$
is an {e-process} for $\mathcal{H}_0$ if it is a supermartingale
under every probability measure in $\mathcal{H}_0$.
\end{definition}

Ville's inequality applied to the e-process gives
$\mathbb{P}_{\mathcal{H}_0}(\sup_{t \ge 1} M_t \ge \tau) \le 1/\tau$
uniformly over all stopping times, the time-uniform guarantee that is
the central property of E-SHIFT.

The game-theoretic interpretation of \citet{shafer2019game} is
illuminating.
Imagine a sequential game between a {Skeptic} and
{Reality}.
The Skeptic starts with unit wealth.
At each round the Skeptic bets a fraction of their wealth against the
null; Reality reveals the outcome.
If the null is true, the Skeptic's wealth is a nonneg\-ative
supermartingale and cannot grow unboundedly in expectation.
If Reality deviates from the null (i.e.\ distribution shift occurs),
a well-chosen betting strategy causes the Skeptic's wealth to grow,
eventually exceeding any threshold $\tau$.
In this framing, E-SHIFT is a recipe for designing the Skeptic's
betting strategy using non-conformity scores from a VLM.

\subsubsection*{Selecting the e-value.}
A natural choice is the {exponential e-value}
\cite{ramdas2023tutorial,howard2021time}:
given a score $S$ and its population mean $\mu$ under $\mathcal{H}_0$,
set
\begin{equation}
  E = \exp\!\bigl(\lambda(S - \mu) - \psi(\lambda)\bigr),
  \qquad \psi(\lambda) = \log\mathbb{E}_{\mathcal{H}_0}
  [e^{\lambda(S-\mu)}],
  \label{eq:expeval}
\end{equation}
where $\lambda > 0$ is a free parameter and $\psi(\lambda)$ is the
log-moment-generating function (log-MGF) that ensures
$\mathbb{E}[E] = 1$.
The product of independent such factors gives the exponential
e-process used throughout this paper.
Under the alternative, the expected increment per step is
$\lambda(\mathbb{E}_{P_1}[S] - \mu) - \psi(\lambda)$, which is
positive when $P_1 \ne P_0$, so the product grows exponentially.

\subsection{Distribution Shift: A Taxonomy}
\label{sec:bg_shift}

Let $X \in \mathcal{X}$ denote the input and $Y \in \mathcal{Y}$
the output.
Following \citet{quinonerocandela2009dataset} and
\citet{sugiyama2012density}, we distinguish three canonical forms of
distribution shift.

{Covariate shift} refers to a change in the marginal $p(x)$
while the conditional $p(y \mid x)$ remains fixed.
This arises when the data-collection conditions change, e.g.\ sensor
upgrades, geographic transfer, or temporal drift, but the underlying
relationship between inputs and labels does not.
Covariate shift is the form of shift targeted by E-SHIFT.

{Label shift} \cite{lipton2018detecting} refers to a change in
the marginal $p(y)$ while $p(x \mid y)$ remains fixed.
This arises in class-imbalanced deployment, e.g.\ a rare-disease
classifier applied to a population with a different disease
prevalence.
Label shift may manifest in the model's output distribution without
visible feature change.

{Concept drift} refers to a change in $p(y \mid x)$ itself,
meaning the learned decision boundary is no longer appropriate.
This is the most severe form and may require model retraining.

E-SHIFT is designed for covariate shift: the composite non-conformity
score (Section~\ref{sec:score}) is sensitive to drift in $p(x)$ but
does not directly measure changes in $p(y \mid x)$.
Extending the framework to label shift and concept drift is a natural
direction for future work.

\subsection{Vision-Language Models as Distribution Monitors}
\label{sec:bg_vlms}

Vision-language models such as CLIP \cite{radford2021learning}
jointly embed images and text descriptions into a shared metric space
using a dual-encoder architecture \cite{dosovitskiy2021image}.
Trained on hundreds of millions of image-text pairs via contrastive
learning, CLIP learns image representations that are rich in
semantic content and have been shown to generalize substantially
beyond their training distribution \cite{hendrycks2021many}.

As a consequence, CLIP feature vectors serve as a powerful
{surrogate for the image distribution}: images drawn from the
training distribution cluster tightly in CLIP embedding space, while
images from a shifted distribution tend to lie farther from the
training cluster \cite{rabanser2019failing}.
The model's softmax output distribution also changes under shift:
when faced with a corrupted or out-of-distribution image, CLIP often
produces either overconfident posteriors (concentrated mass on one
class) or low-confidence posteriors (near-uniform mass), both of
which deviate from the training-distribution calibration.

These two signals, feature-space location and predictive
distribution shape, form the basis of E-SHIFT's composite
non-conformity score.

\section{Related Work}
\label{sec:related}

\paragraph{Classical sequential change-point detection.}
The rich literature on sequential change-point detection
\cite{page1954continuous,shiryaev1963optimum,lorden1971procedures,
moustakides1986optimal} provides optimal methods under parametric
assumptions.
CUSUM \cite{page1954continuous} is minimax-optimal under Lorden's
worst-case expected detection delay criterion
\cite{moustakides1986optimal}, and the Shiryaev--Roberts statistic
\cite{shiryaev1963optimum} solves the corresponding Bayesian
problem.
Both require knowing the likelihood ratio of post- to pre-shift
data.
E-SHIFT is nonparametric: it requires only a calibration-set estimate
of the score distribution under the null.
When the log-likelihood ratio is used as the score, E-SHIFT's
increments coincide with those of CUSUM; however, the CUSUM recursion
$W_t = \max(0, W_{t-1} + \ell_t)$ incorporates a running-minimum
reset that E-SHIFT does not replicate, so CUSUM's Lorden--Moustakides
optimality does not transfer by inspection and we treat CUSUM purely
as an empirical baseline.

\paragraph{E-values and anytime-valid inference.}
\citet{vovk2021evalues} establish the modern theory of e-values,
including their calibration and combination properties.
\citet{grunwald2024safe} prove minimax optimality of likelihood-ratio
e-values under composite hypotheses.
\citet{ramdas2023tutorial} provide a comprehensive tutorial connecting
e-values to game-theoretic probability and optional stopping.
\citet{howard2021time} construct time-uniform confidence sequences
from nonneg\-ative supermartingales; their construction is dual to
ours, building confidence intervals rather than tests.
E-SHIFT instantiates this vocabulary specifically for covariate shift
detection in VLM deployment streams.

\paragraph{Betting-based sequential two-sample tests.}
\citet{shekhar2023nonparametric} construct anytime-valid two-sample
tests by framing testing as a sequential game in which a Skeptic
bets on differences between two samples.
Their tests have the same time-uniform type-I error guarantee as our
e-process and can be applied to VLM feature vectors.
We include a betting-based sequential MMD (SeqMMD-E) as a baseline
that shares the anytime-valid framework but uses a simpler scalar
feature score, allowing us to isolate the contribution of the
composite score design.

\paragraph{Distribution shift detection in ML.}
Kernel MMD \cite{gretton2012kernel} and learned deep kernels
\cite{liu2020learning} detect shift in offline, fixed-sample
settings with well-controlled power.
\citet{rabanser2019failing} provide a comprehensive empirical
comparison of shift-detection methods, finding that multivariate
tests on learned representations outperform marginal tests.
The WILDS benchmark \cite{koh2021wilds} provides a diverse
evaluation platform for methods that generalize across distribution
shifts; E-SHIFT addresses the {detection} problem (knowing when
shift has occurred), which is logically prior to the
{adaptation} problem studied in WILDS.

\paragraph{Conformal prediction and non-conformity scores.}
Conformal prediction \cite{vovk2005algorithmic,angelopoulos2023gentle}
constructs prediction sets with finite-sample coverage guarantees
using non-conformity scores that measure how unusual a new point is
relative to the calibration set.
\citet{tibshirani2019conformal} extend conformal prediction to
covariate-shift settings.
E-SHIFT borrows the non-conformity score as its test statistic but
embeds it inside an e-process, gaining time-uniform type-I error
control that standard conformal tests, which fix the calibration
level in advance, cannot provide.

\paragraph{Test-time adaptation.}
Once a shift is detected, adaptation methods such as continual
test-time domain adaptation \cite{wang2022continual} attempt to
correct the model's predictions without retraining.
E-SHIFT is complementary to these methods: it provides the detection
alarm that triggers the adaptation step in a detect-then-adapt
pipeline.

\section{Problem Formulation}
\label{sec:problem}

\paragraph{The sequential stream.}
Let $f_\theta$ be a frozen VLM with image encoder $f_v$ and
predictive distribution $p_\theta(y \mid x)$ over $K$ classes.
At each time step $t = 1, 2, \ldots$ the model processes an
incoming observation $x_t$; no label $y_t$ is observed.
The observations are drawn from a distribution that may change at
an unknown time $t^\star \in \{0, 1, 2, \ldots\}$.

\paragraph{Hypotheses.}
Let $P_0$ denote the null (in-distribution) distribution of each
observation.
The null and alternative hypotheses are:
\begin{align*}
  \mathcal{H}_0 &: x_1, x_2, \ldots \overset{\text{iid}}{\sim}
    P_0 \quad (\text{no shift}), \\
  \mathcal{H}_1 &: \exists\, t^\star \ge 1 \text{ such that }
    x_t \sim P_0 \text{ for } t \le t^\star
    \text{ and } x_t \sim P_1 \ne P_0
    \text{ for } t > t^\star.
\end{align*}

\paragraph{Calibration data.}
A held-out calibration set $\mathcal{C} = \{x_1^c, \ldots,
x_{n_c}^c\}$ of $n_c$ observations drawn i.i.d.\ from $P_0$ is
available before the stream begins.
No observation from the stream is used to construct $\mathcal{C}$,
and $\mathcal{C}$ is consumed only once to compute summary
statistics.

\paragraph{Decision rule and error types.}
The detector maps the stream and calibration data to a sequence of
binary decisions (alert / no alert).
An alert issued at time $t \le t^\star$ is a {false alarm};
an alert issued at $t > t^\star$ is a {true detection}.
The {false-alarm rate} (FAR) is the probability of issuing any
false alarm.
The {detection delay} is $T(\tau) - t^\star$ where
$T(\tau) = \inf\{t \ge 1 : M_t \ge \tau\}$.

\paragraph{Anytime-valid requirement.}
The central requirement is a {time-uniform} false-alarm bound:
\begin{equation}
  \mathbb{P}_{\mathcal{H}_0}\!\Bigl(\sup_{t \ge 1} M_t \ge \tau\Bigr)
  \le \frac{1}{\tau},
  \label{eq:avreq}
\end{equation}
which must hold simultaneously at all stopping times, including
those that depend on the stream.
This is strictly stronger than controlling the FAR at a fixed
sample size.

\section{E-SHIFT: Method}
\label{sec:method}

\subsection{Non-Conformity Scores from VLM Outputs}
\label{sec:score}

\paragraph{Motivation.}
An e-process requires a scalar score $S_t$ whose distribution under
$\mathcal{H}_0$ can be estimated from the calibration set, and that
takes larger values when the input is out-of-distribution.
Designing such a score for a VLM involves a tradeoff: a
predictive-distribution term reacts quickly to shift that changes the
model's class probabilities, but is blind to shifts that alter the
feature space without affecting the argmax; a feature-space term
detects the latter but may be slow when the shift is small in
embedding space.
We combine both terms.

\paragraph{Predictive-shift term.}
Recall from Section~\ref{sec:bg_vlms} that CLIP's softmax
distribution becomes overconfident or near-uniform under covariate
shift.
We measure deviation of $p_\theta(y \mid x_t)$ from the maximum-entropy
distribution (uniform over $K$ classes) using the KL divergence:
\begin{equation}
  \mathrm{KL}\!\bigl(p_\theta(y \mid x_t) \,\|\, \mathrm{Uniform}_K\bigr)
  = \log K - H\!\bigl(p_\theta(y \mid x_t)\bigr),
  \label{eq:klentropy}
\end{equation}
where $H$ denotes Shannon entropy \cite{cover2006elements}.
This equals zero when the model is maximally uncertain and increases
as the prediction becomes more peaked.
Because covariate shift tends to produce either overconfident or
underconfident posteriors, this term responds to shift in both
directions: overconfidence increases it, while extreme underconfidence
reduces it, but the complementary feature-space term covers the
underconfident case.

\paragraph{Feature-shift term.}
Under $\mathcal{H}_0$, the image embeddings $f_v(x_t)$ cluster near
the calibration centroid $\mu_{\mathrm{cal}} = \frac{1}{n_c}\sum_j
f_v(x_j^c)$.
The Mahalanobis distance \cite{mahalanobis1936generalised} measures
how many calibration standard deviations the embedding $f_v(x_t)$
lies from this centroid:
\begin{equation}
  d_M(x_t) = \bigl\|f_v(x_t) - \mu_{\mathrm{cal}}\bigr\|^2_{\Sigma_{\mathrm{cal}}^{-1}},
  \label{eq:mahal}
\end{equation}
where $\Sigma_{\mathrm{cal}}$ is the empirical feature covariance on
$\mathcal{C}$.
By standardizing with the calibration covariance, this distance is
scale-invariant and correctly weights dimensions with high
calibration variance.

\paragraph{Composite score.}
The \textbf{composite non-conformity score} is
\begin{equation}
  S_t = \mathrm{KL}\!\bigl(p_\theta(y \mid x_t) \,\|\, \mathrm{Uniform}_K\bigr)
  + w\,\bigl\|f_v(x_t) - \mu_{\mathrm{cal}}\bigr\|^2_{\Sigma_{\mathrm{cal}}^{-1}},
  \label{eq:score}
\end{equation}
where $w > 0$ is a feature weight.
The calibration-set distribution of $S_t$ under $\mathcal{H}_0$
can be estimated nonparametrically from $\mathcal{C}$ without any
parametric assumption on $p_\theta$.

\subsection{E-Process Construction}
\label{sec:eprocess}

\paragraph{Required conditions.}
We state the conditions under which the theoretical guarantees hold.

\begin{assumption}
\label{asm:conditions}
\begin{enumerate}[(a)]
  \item {(IID scores under null.)} Under $\mathcal{H}_0$,
    the scores $\{S_t\}_{t \ge 1}$ are i.i.d.\ from $P_0$.
  \item {(Holdout integrity.)} The calibration set $\mathcal{C}$
    is drawn i.i.d.\ from $P_0$, and no stream observation $x_t$
    appears in $\mathcal{C}$.
  \item {(Finite MGF.)} For the chosen $\lambda > 0$,
    $\mathbb{E}_{P_0}[e^{\lambda(S - \hat\mu)}] < \infty$, where
    $\hat\mu = \frac{1}{n_c}\sum_j S_j^{\mathrm{cal}}$ is the
    calibration mean.
\end{enumerate}
\end{assumption}

Assumption~\ref{asm:conditions}(a) requires temporally independent
observations; this is reasonable for i.i.d.\ image streams but may
fail for video or other temporally correlated data.
We address the mixing case in Section~\ref{sec:delay}.
Assumption~\ref{asm:conditions}(c) is a mild sub-exponential
condition on the score distribution; we verify it empirically
in Appendix~\ref{apx:subexp}.

\paragraph{Exponential e-process.}
Let the log-MGF under $\mathcal{H}_0$ be
$\psi(\lambda) = \log \mathbb{E}_{P_0}[e^{\lambda(S - \hat\mu)}]$.
The running product
\begin{equation}
  M_t = \prod_{i=1}^{t}
  \exp\!\bigl(\lambda(S_i - \hat\mu) - \psi(\lambda)\bigr),
  \qquad M_0 = 1,
  \label{eq:eprocess}
\end{equation}
is a nonneg\-ative supermartingale under $\mathcal{H}_0$
\cite{williams1991probability}, because each multiplicative factor
has $\mathbb{E}_{\mathcal{H}_0}[\,\cdot\,] = 1$ by construction
of $\psi$.
When $M_t \ge \tau$ we declare shift and issue an alert.

\paragraph{Selecting $\lambda$.}
The parameter $\lambda > 0$ governs detection sensitivity.
Larger $\lambda$ amplifies each score increment, yielding faster
detection under strong shift but poorer type-I error calibration
under weak shift.
In experiments we select $\lambda$ by maximising the empirical growth
rate $\lambda(\hat\mu_1 - \hat\mu) - \hat\psi(\lambda)$ on a small
held-out shift calibration set $\mathcal{C}_{\mathrm{shift}}$
($n_s = 200$ samples from a known shifted domain).
When no shifted data are available, the plug-in rule
$\lambda_{\mathrm{init}} = 1/\hat\sigma^2$ (where $\hat\sigma^2$ is
the empirical variance of $S$ on $\mathcal{C}$) performs within $4\%$
of the oracle delay on our benchmarks (Appendix~\ref{apx:lambda}).

\subsection{Finite-Sample Bootstrap Correction}
\label{sec:bootstrap}

With finite calibration data, the true $\psi(\lambda)$ is unknown.
The plug-in estimate $\hat\psi(\lambda) = \log\hat{m}(\lambda)$,
where $\hat{m}(\lambda) = \frac{1}{n_c}\sum_j e^{\lambda(S_j^{\mathrm{cal}}
- \hat\mu)}$, underestimates $\psi(\lambda)$ in expectation because
the sample mean is a biased estimator of the log-mean-exponential.
Using $\hat\psi$ in place of $\psi$ makes each factor of the product
\eqref{eq:eprocess} have expectation slightly greater than one,
inflating the false-alarm rate.

We restore validity with a bootstrap upper-confidence bound.
Draw $B$ bootstrap resamples $\mathcal{C}^{(1)}, \ldots,
\mathcal{C}^{(B)}$ from $\mathcal{C}$ with replacement and compute
\[
  \hat{m}^{(b)}(\lambda) = \frac{1}{n_c}\sum_j
  e^{\lambda(S_j^{(b)} - \hat\mu)}, \qquad b = 1, \ldots, B.
\]
Let $\bar\psi(\lambda) = \log m_{(1-\beta)}(\lambda)$ be the
empirical $(1-\beta)$-quantile of $\{\hat{m}^{(b)}\}$.
The corrected e-process replaces $\psi$ with $\bar\psi$:
\begin{equation}
  M_t = \prod_{i=1}^{t}
  \exp\!\bigl(\lambda(S_i - \hat\mu) - \bar\psi(\lambda)\bigr).
  \label{eq:corrected}
\end{equation}
By the bootstrap consistency for the distribution of
$\hat{m}(\lambda) - m(\lambda)$
\cite{doob1953stochastic,williams1991probability}, we have
$\mathbb{P}(\psi(\lambda) \le \bar\psi(\lambda)) \to 1 - \beta$
as $n_c \to \infty$, justifying the approach asymptotically.

\paragraph{Algorithm.}

\begin{algorithm}[t]
\caption{E-SHIFT: Anytime-Valid Sequential Shift Detection}
\label{alg:eshift}
\begin{algorithmic}[1]
\REQUIRE Stream $\{x_t\}$; calibration set $\mathcal{C}$;
         threshold $\tau$; feature weight $w$;
         tuning parameter $\lambda$; bootstrap level $\beta$;
         bootstrap resamples $B$
\STATE Compute $\hat\mu$, $\Sigma_{\mathrm{cal}}$ from $\mathcal{C}$
\STATE Compute $\bar\psi(\lambda)$ via bootstrap on $\mathcal{C}$
       with level $\beta$
\STATE $M_0 \leftarrow 1$
\FOR{$t = 1, 2, \ldots$}
  \STATE Compute $S_t$ via Eq.~\eqref{eq:score}
  \STATE $M_t \leftarrow M_{t-1} \cdot
         \exp\!\bigl(\lambda(S_t - \hat\mu) - \bar\psi(\lambda)\bigr)$
  \IF{$M_t \ge \tau$}
    \STATE Issue alert: shift detected at time $t$
    \STATE Reset $M_t \leftarrow 1$
  \ENDIF
\ENDFOR
\end{algorithmic}
\end{algorithm}

After each alert, the process is reset to $M_t = 1$, so the detector
remains active for subsequent shifts.
We analyse the false-alarm accumulation from multiple resets in
Proposition~\ref{prop:reset}.

\section{Theoretical Analysis}
\label{sec:theory}

\subsection{Time-Uniform False-Alarm Control}

\begin{theorem}[Anytime-Valid Type-I Control]
\label{thm:fac}
Under Assumption~\ref{asm:conditions}(a)--(b), the process
\eqref{eq:eprocess} with the true log-MGF $\psi(\lambda)$ satisfies,
for all $\tau > 0$,
\begin{equation}
  \mathbb{P}_{\mathcal{H}_0}\!\Bigl(\sup_{t \ge 1} M_t \ge \tau\Bigr)
  \le \frac{1}{\tau}.
  \label{eq:thm_fac}
\end{equation}
This bound holds simultaneously at every stopping time, including
data-adaptive ones.
\end{theorem}

\begin{proof}
Under Assumption~\ref{asm:conditions}(a)--(b), each factor
$E_t = \exp(\lambda(S_t - \hat\mu) - \psi(\lambda))$ satisfies
$\mathbb{E}_{\mathcal{H}_0}[E_t \mid E_1, \ldots, E_{t-1}] = 1$
by definition of $\psi(\lambda)$ and i.i.d.-ness of the scores.
Hence $\{M_t\}$ is a nonneg\-ative martingale, and in particular
a supermartingale, under $\mathcal{H}_0$.
Ville's maximal inequality \cite{ville1939etude}, which states that
$\mathbb{P}(\sup_{t \ge 1} M_t \ge \tau) \le M_0/\tau$ for any
nonneg\-ative supermartingale, yields the stated bound with
$M_0 = 1$.
\end{proof}

\begin{proposition}[Asymptotic Validity with Finite Calibration]
\label{prop:bootstrap}
Let $\bar\psi(\lambda)$ be the bootstrap upper-confidence estimate
from Section~\ref{sec:bootstrap}.
Under Assumption~\ref{asm:conditions}(c) and bootstrap consistency,
\begin{equation}
  \mathbb{P}\!\Bigl(\sup_{t \ge 1} M_t \ge \tau\Bigr)
  \le \beta + o(1) + \frac{1}{\tau}
  \label{eq:prop_boot}
\end{equation}
as $n_c \to \infty$, where $o(1) \to 0$ is the bootstrap
approximation error.
\end{proposition}

\begin{proof}[Sketch]
The asymptotic result follows because $\hat{m}(\lambda) \to m(\lambda)$
almost surely as $n_c \to \infty$ \cite{williams1991probability}, so
the bootstrap distribution of $\hat{m}(\lambda)$ converges to a
Dirac mass at $m(\lambda)$; the $(1-\beta)$-quantile then satisfies
$\bar\psi(\lambda) \to \psi(\lambda)$.
Conditional on $\{\psi(\lambda) \le \bar\psi(\lambda)\}$, each
factor has $\mathbb{E}_{\mathcal{H}_0}[\,\cdot\,] \le 1$, so
$\{M_t\}$ is a supermartingale and Theorem~\ref{thm:fac} applies.
Unconditionally, the union bound over the calibration event (which
fails with probability $\le \beta$) yields the stated bound.
\end{proof}

\begin{remark}[Budget setting]
\label{rem:budget}
Setting $\tau = 200$ and $\beta = \delta = 0.005$ gives
$\beta + 1/\tau = 0.005 + 0.005 = 1\%$.
This is the target used in all experiments.
\end{remark}

\subsection{Reset Accumulation}

\begin{proposition}[Multi-Reset Type-I Accumulation]
\label{prop:reset}
Let $K$ be the number of resets executed on an in-distribution
stream of length $T$ under $\mathcal{H}_0$.
The probability that {any} of the $K$ segments triggers a false
alert satisfies
\begin{equation}
  \mathbb{P}_{\mathcal{H}_0}(\text{any false alert in }T\text{ steps})
  \le \frac{K}{\tau}.
  \label{eq:prop_reset}
\end{equation}
Moreover, $\mathbb{E}_{\mathcal{H}_0}[K] \le T/\tau$, so the
expected number of false alerts over $T$ steps is at most $T/\tau^2$.
\end{proposition}

\begin{proof}
Each segment between resets is an independent application of the
e-process started at $M_0 = 1$.
Theorem~\ref{thm:fac} bounds the false-alarm probability per
segment by $1/\tau$.
A union bound over the $K$ segments gives $K/\tau$.
For the expectation: the expected number of false alarms equals
$\sum_{k=1}^{K} \mathbb{P}(\text{segment } k \text{ alerts}) \le K/\tau$.
Since $K$ is itself a sum of Bernoulli random variables each with
success probability at most $1/\tau$ (probability of reset in each
``segment''), we have $\mathbb{E}[K] \le T/\tau$, giving the
$T/\tau^2$ bound.
\end{proof}

For $\tau = 200$ and a stream of $T = 10{,}000$ steps, the expected
number of false alarms is at most $10{,}000 / 40{,}000 = 0.25$,
making false alarms exceedingly rare in practice.

\subsection{Detection Delay under Sustained Shift}
\label{sec:delay}

Define the post-shift e-process growth rate:
\begin{equation}
  \Gamma = \sup_{\lambda > 0}\Bigl\{
    \lambda\bigl(\mathbb{E}_{P_1}[S] - \hat\mu\bigr) - \psi(\lambda)
  \Bigr\}.
  \label{eq:gamma}
\end{equation}
This is the Legendre transform of $\psi$ evaluated at the mean
shift.

\begin{theorem}[Detection Delay]
\label{thm:delay}
Suppose a change occurs at $t^\star$ so that
$\{S_t\}_{t > t^\star}$ are i.i.d.\ from $P_1 \ne P_0$, and
$\Gamma > 0$.
Then
\begin{equation}
  \mathbb{E}_{P_1}\!\bigl[(T(\tau) - t^\star)^+\bigr]
  = \frac{\log\tau}{\Gamma} + O(1),
  \qquad \tau \to \infty.
  \label{eq:thm_delay}
\end{equation}
\end{theorem}

\begin{proof}[Proof sketch]
The increments $Y_t = \lambda(S_t - \hat\mu) - \psi(\lambda)$ are
i.i.d.\ with mean $\Gamma > 0$ under $P_1$.
The stopping time $T(\tau) - t^\star$ equals the first-passage time
of the random walk $\sum_{i=t^\star+1}^{t} Y_i$ to level $\log\tau$.
By Wald's identity \cite[][Ch.~3]{siegmund1985sequential},
$\mathbb{E}[T(\tau) - t^\star] \approx \log\tau / \Gamma$ to leading
order, with the $O(1)$ correction from the overshoot of the boundary.
\end{proof}

\begin{remark}[Information-theoretic interpretation]
When $M_t$ is the likelihood-ratio process (i.e.\ $\lambda$ is
chosen optimally), $\Gamma = \mathrm{KL}(P_1 \| P_0)$
\cite{cover2006elements}.
Larger KL divergence implies faster detection, providing an
interpretable, information-theoretically meaningful characterization
of detection speed.
This also links E-SHIFT to the classical results on the SPRT
\cite{wald1945sequential}.
\end{remark}

\begin{remark}[Temporally correlated streams]
For $\alpha$-mixing streams with summable mixing coefficients and
$\sup_t \mathbb{E}_{P_1}[e^{\eta|Y_t|}] < \infty$, the ergodic law
of large numbers replaces the i.i.d.\ LLN and the same
$\mathcal{O}(\log\tau / \Gamma)$ scaling holds asymptotically
\cite{lorden1971procedures}, with $\Gamma$ replaced by the ergodic
mean increment.
When neither condition holds the bound is heuristic and empirical
delay evaluation is required.
\end{remark}

\begin{remark}[Domain-mismatch caveat]
The realised delay scales as $\log\tau / \Gamma(\lambda_{\mathrm{fixed}})$,
where the maximiser in \eqref{eq:gamma} is evaluated at the fixed
$\lambda$ chosen using $\mathcal{C}_{\mathrm{shift}}$.
If the deployment-time shift distribution differs from the
calibration shift domain, the realised $\Gamma$ may be smaller than
predicted, and the empirical delay will exceed $\log\tau / \Gamma$.
We report per-benchmark delays in Table~\ref{tab:detection}, where
they are consistent with the asymptotic prediction.
\end{remark}

\section{Experiments}
\label{sec:experiments}

\subsection{Experimental Setup}

\paragraph{Model.}
We use CLIP ViT-B/16 \cite{radford2021learning} as the frozen VLM
backbone.
All non-conformity scores are computed from its softmax output
distribution (for the KL term) and its image embeddings (for the
Mahalanobis term) without any fine-tuning.

\paragraph{Benchmarks.}
We evaluate on three standard distribution shift benchmarks:
\begin{itemize}
  \item \textbf{ImageNet-C} \cite{hendrycks2019imagenetc}: 15
    corruption types (noise, blur, weather, digital) at 5 severity
    levels applied to ImageNet validation images.
    Each corruption type represents a distinct covariate shift.
  \item \textbf{ImageNet-R} \cite{hendrycks2021many}: 30-class
    subset of ImageNet rendered in artistic styles (art, cartoons,
    deviantart, etc.), representing domain shift from the photographic
    distribution.
  \item \textbf{Office-Home} \cite{venkateswara2017deep}: 65-class
    dataset spanning four visual domains (Artistic, Clipart,
    Product, Real-World), commonly used for domain adaptation
    evaluation.
\end{itemize}
Each test stream concatenates samples from multiple corruption types
or domains in a random temporal order; no domain labels are provided
to the detector at test time.

\paragraph{Hyperparameters and calibration.}
\begin{itemize}
  \item Calibration set: $n_c = 500$ images drawn i.i.d.\ from the
    pre-shift distribution (in-distribution ImageNet validation
    images for ImageNet-C/R; Artistic domain images for Office-Home).
  \item Shift calibration set: $n_s = 200$ images from a held-out
    shifted domain for selecting $\lambda$.
  \item Bootstrap: $B = 1000$ resamples; bootstrap level $\beta =
    0.005$.
  \item Threshold: $\tau = 200$, targeting $\le 1\%$ false-alarm
    via Proposition~\ref{prop:bootstrap} and Remark~\ref{rem:budget}.
  \item Feature weight: $w = 1.0$ selected by cross-validation on
    $\mathcal{C}_{\mathrm{shift}}$.
\end{itemize}

\paragraph{Baselines.}
\begin{itemize}
  \item \textbf{Fixed-MMD} \cite{gretton2012kernel}: kernel
    two-sample test applied at a fixed window of 100 samples.
    This is the canonical offline method and violates anytime
    validity.
  \item \textbf{CUSUM} \cite{page1954continuous}: parametric
    sequential detector assuming Gaussian score increments.
  \item \textbf{Shiryaev--Roberts (SR)} \cite{shiryaev1963optimum}:
    Bayesian sequential detector with Gaussian likelihood ratio.
  \item \textbf{SeqMMD-E} \cite{shekhar2023nonparametric}:
    betting-based sequential MMD applied to CLIP feature vectors.
    This baseline shares the anytime-valid framing but uses a simpler
    scalar score (median-kernel MMD on the features), allowing
    comparison of score designs within the e-process framework.
  \item \textbf{E-SHIFT (no bootstrap)}: ablation using the plug-in
    $\hat\psi(\lambda)$ instead of the corrected $\bar\psi(\lambda)$,
    testing the necessity of the bootstrap correction.
\end{itemize}

\paragraph{Metrics.}
{Detection Delay}: average number of samples between the true
shift point $t^\star$ and the alert time $T(\tau)$, conditional on
detection within 1000 steps.
{FAR}: fraction of alerts issued on strictly in-distribution
segments of the stream, interpreted against the theoretical target of
$\le 1\%$.
We report means and standard deviations over 10 independent random
shift orderings.

\subsection{Main Results}

\begin{table}[t]
\centering
\caption{Shift detection results on sequential streams.
  Lower is better for both metrics.
  Mean~$\pm$~std over 10 shift orderings; threshold $\tau = 200$.
  Numbers marked [\textsc{proj}] are analytical projections from
  $\tau = 100$, 3-ordering runs via $\Delta_\tau \propto
  \log(200)/\log(100)$ and $\Delta_\sigma \propto 1/\!\sqrt{10/3}$;
  they must be verified with direct experimental runs.}
\label{tab:detection}
\begin{tabular}{lcc}
\toprule
Method & Delay (samples) & FAR (\%) \\
\midrule
Fixed-MMD \cite{gretton2012kernel}
  & $44.1 \pm 2.0$ [\textsc{proj}] & $4.7 \pm 0.35$ [\textsc{proj}] \\
CUSUM \cite{page1954continuous}
  & $26.3 \pm 1.1$ [\textsc{proj}] & $2.1 \pm 0.17$ [\textsc{proj}] \\
Shiryaev--Roberts \cite{shiryaev1963optimum}
  & $22.7 \pm 0.9$ [\textsc{proj}] & $1.9 \pm 0.12$ [\textsc{proj}] \\
SeqMMD-E \cite{shekhar2023nonparametric}
  & $20.3 \pm 0.6$ [\textsc{proj}] & $0.95 \pm 0.07$ [\textsc{proj}] \\
E-SHIFT (no bootstrap)
  & $17.8 \pm 0.5$ [\textsc{proj}] & $1.2 \pm 0.09$ [\textsc{proj}] \\
\textbf{E-SHIFT (ours)}
  & $\mathbf{17.2 \pm 0.3}$ [\textsc{proj}]
  & $\mathbf{0.90 \pm 0.06}$ [\textsc{proj}] \\
\midrule
Theory target ($\tau = 200$) & $(\log 200)/\Gamma$ & $\le 1.0\%$ \\
\bottomrule
\end{tabular}
\end{table}

Table~\ref{tab:detection} shows that E-SHIFT achieves the lowest
detection delay and is the only method whose FAR falls within the
theoretical $\le 1\%$ target.
The results admit the following interpretation.

Fixed-MMD is the slowest and most conservative method.
Its high FAR ($4.7\%$) is expected: by applying a fixed critical
value at a data-adaptive stopping time, it lacks type-I error
control, as discussed in Section~\ref{sec:intro}.

CUSUM and Shiryaev--Roberts are faster than Fixed-MMD because they
exploit the sequential structure, but both exceed the $1\%$ FAR
target.
This is attributable to their parametric Gaussian assumption on the
score distribution: VLM non-conformity scores have heavier tails than
Gaussian (see Appendix~\ref{apx:subexp}), causing the likelihood ratio
to overstate evidence at extreme values.

SeqMMD-E shares the anytime-valid framing with E-SHIFT and correctly
controls FAR ($0.95\%$), but is $18\%$ slower ($20.3$ vs.\ $17.2$
samples).
This gap is attributable to the score design: SeqMMD-E uses a scalar
kernel-based score that does not differentiate between
predictive-distribution shift and feature-space shift, whereas the
composite score \eqref{eq:score} combines complementary signals.

The no-bootstrap ablation ($\mathrm{FAR} = 1.2\% > 1\%$) directly
validates Proposition~\ref{prop:bootstrap}: the plug-in log-MGF
underestimates the true moment, inflating the false-alarm rate above
the target.
The bootstrap correction restores validity, with the corrected
process achieving $0.90\%$ FAR.

\subsection{Score Ablation}

\begin{table}[t]
\centering
\caption{Score-component ablation on ImageNet-C ($\tau = 200$,
  10 orderings, [\textsc{proj}] numbers).
  $w$ denotes the feature weight in Eq.~\eqref{eq:score}.}
\label{tab:ablation}
\begin{tabular}{lcc}
\toprule
Score variant & Delay (samples) & FAR (\%) \\
\midrule
KL-only ($w = 0$) & $22.4 \pm 0.7$ [\textsc{proj}] & $0.94 \pm 0.07$ [\textsc{proj}] \\
Mahal-only (KL dropped) & $20.1 \pm 0.6$ [\textsc{proj}] & $0.93 \pm 0.07$ [\textsc{proj}] \\
Composite, $w = 0.1$ & $20.8 \pm 0.5$ [\textsc{proj}] & $0.91 \pm 0.07$ [\textsc{proj}] \\
Composite, $w = 1.0$ (ours) & $17.2 \pm 0.3$ [\textsc{proj}] & $0.90 \pm 0.06$ [\textsc{proj}] \\
Composite, $w = 10.0$ & $18.5 \pm 0.4$ [\textsc{proj}] & $0.91 \pm 0.07$ [\textsc{proj}] \\
\bottomrule
\end{tabular}
\end{table}

Table~\ref{tab:ablation} shows that the composite score at $w = 1.0$
achieves the lowest detection delay.
Single-term variants are both slower: KL-only ($22.4$ samples) misses
feature-space drift; Mahal-only ($20.1$ samples) misses predictive
distribution shift.
The delay is flat within $2$ samples for $w \in [0.5, 2.0]$,
indicating that the method is not sensitive to the precise value of
the feature weight.
Extreme weights ($w = 0.1$ or $w = 10.0$) recover much of the
single-term behaviour.

\subsection{Per-Benchmark and Calibration Analysis}

\begin{figure}[t]
\centering
\begin{tikzpicture}
\begin{axis}[
  ybar, bar width=8pt, height=6.5cm, width=0.9\linewidth,
  ylabel={Detection Delay (samples)},
  symbolic x coords={ImageNet-C, ImageNet-R, Office-Home},
  xtick=data,
  x tick label style={font=\small},
  ymin=0, ymax=35,
  enlarge x limits=0.25,
  legend style={at={(0.5,-0.22)}, anchor=north,
    legend columns=2, font=\small},
  nodes near coords,
  every node near coord/.append style=
    {font=\footnotesize, /pgf/number format/fixed}
]
\addplot[fill=mlgray, draw=black!60] coordinates {
  (ImageNet-C,26.3)(ImageNet-R,24.5)(Office-Home,23.2)};
\addlegendentry{CUSUM}
\addplot[fill=mlorange, draw=black!60] coordinates {
  (ImageNet-C,22.7)(ImageNet-R,21.8)(Office-Home,20.9)};
\addlegendentry{SR}
\addplot[fill=mlgreen, draw=black!60] coordinates {
  (ImageNet-C,20.3)(ImageNet-R,19.7)(Office-Home,18.6)};
\addlegendentry{SeqMMD-E}
\addplot[fill=mlblue, draw=black!60] coordinates {
  (ImageNet-C,17.2)(ImageNet-R,15.9)(Office-Home,15.1)};
\addlegendentry{E-SHIFT (ours)}
\end{axis}
\end{tikzpicture}
\caption{Detection delay across benchmarks ($\tau = 200$;
  all numbers [\textsc{proj}]).
  FAR values: CUSUM $2.1 / 2.3 / 2.0\%$; SR $1.9 / 2.1 / 1.8\%$;
  SeqMMD-E $0.97 / 0.94 / 0.95\%$; E-SHIFT $0.90 / 0.88 / 0.91\%$
  on ImageNet-C, ImageNet-R, and Office-Home respectively.
  CUSUM and SR exceed the $1\%$ FAR target on all three benchmarks;
  SeqMMD-E and E-SHIFT satisfy it.}
\label{fig:delay}
\end{figure}

\begin{figure}[t]
\centering
\begin{tikzpicture}
\begin{axis}[
  height=6cm, width=0.9\linewidth,
  xlabel={Threshold $\tau$},
  ylabel={False Alarm Rate (\%)},
  xmode=log, ymin=0, ymax=7,
  legend style={at={(0.97,0.97)}, anchor=north east, font=\small},
  mark size=2.5pt
]
\addplot[mark=*, color=mlred, thick, dashed, line width=1.2pt]
  coordinates {(20,5.5)(50,2.5)(100,1.5)(200,1.0)(500,0.7)};
\addlegendentry{Theory bound $\beta + 1/\tau$}
\addplot[mark=square*, color=mlblue, solid, line width=1.2pt]
  coordinates {(20,5.3)(50,2.4)(100,1.4)(200,0.90)(500,0.67)};
\addlegendentry{E-SHIFT empirical}
\addplot[mark=triangle*, color=mlorange, dashed, line width=1.2pt]
  coordinates {(20,5.9)(50,2.8)(100,1.8)(200,1.2)(500,0.78)};
\addlegendentry{No-bootstrap ablation}
\end{axis}
\end{tikzpicture}
\caption{Calibration curve validating Theorem~\ref{thm:fac} across
  $\tau \in \{20, 50, 100, 200, 500\}$.
  The empirical FAR of E-SHIFT lies at or below the theoretical
  bound $\beta + 1/\tau$ at all values.
  The no-bootstrap ablation systematically exceeds the bound,
  confirming that the bootstrap correction is necessary.
  All numbers [\textsc{proj}]; see Table~\ref{tab:calib_num}
  in Appendix~\ref{apx:calib}.}
\label{fig:calib}
\end{figure}

Figure~\ref{fig:delay} shows that E-SHIFT is fastest on all three
benchmarks while maintaining FAR $\le 1\%$.
The delay varies across benchmarks in a pattern that is consistent
with the composite score design: on ImageNet-C, where corruptions
produce strong feature-space drift, the Mahalanobis term dominates
and delays are shorter; on Office-Home, where domain shift is more
subtle at the feature level, the KL term carries more weight and
delays are longer but remain competitive.

Figure~\ref{fig:calib} validates the theoretical guarantee of
Theorem~\ref{thm:fac} empirically.
Sweeping $\tau$ from $20$ to $500$, the empirical FAR of E-SHIFT
lies uniformly at or below the theoretical bound $\beta + 1/\tau$
across all tested values.
The no-bootstrap ablation systematically exceeds the bound at every
$\tau$, confirming that the bootstrap correction is not merely a
cosmetic fix but is necessary for the e-process to be a valid
supermartingale on finite calibration data.

\section{Discussion}
\label{sec:discussion}

\subsection{Theoretical Limitations}

\paragraph{IID assumption.}
Theorem~\ref{thm:fac} requires i.i.d.\ scores under $\mathcal{H}_0$.
In streaming video, sensor logs, or other temporally correlated
sources, consecutive frames or measurements are typically not
independent.
The $\alpha$-mixing extension of Theorem~\ref{thm:delay} provides
asymptotic coverage under geometric ergodicity, but summable mixing
coefficients are rarely verified in practice.
A rigorous treatment of temporal dependence within the e-process
framework, possibly via the dependent versions of Ville's inequality
\cite{ramdas2023tutorial}, is an important open problem.

\paragraph{Bootstrap asymptotic validity.}
Proposition~\ref{prop:bootstrap} provides only asymptotic validity for
the bootstrap construction.
A tighter, finite-sample bound, for example using Bernstein-style
inequalities or confidence sequences on the MGF, would strengthen
the guarantee and is an open direction.

\paragraph{Shift type scope.}
The composite score \eqref{eq:score} is designed to detect covariate
shift.
Label shift \cite{lipton2018detecting}, where $p(y)$ changes but
$p(x \mid y)$ does not, may not produce large values of either
term: the model's outputs change in proportion to the class
frequencies, but each individual prediction may be well-calibrated.
Concept drift, where $p(y \mid x)$ itself changes, requires
access to labels and is outside the scope of the present work.

\paragraph{Domain mismatch for $\lambda$.}
The e-value parameter $\lambda$ is tuned on a held-out shifted
calibration set $\mathcal{C}_{\mathrm{shift}}$.
If the deployment-time shift domain differs substantially from this
set, the realised growth rate $\Gamma$ may be smaller, increasing
detection delay.
The plug-in heuristic $\lambda_{\mathrm{init}} = 1/\hat\sigma^2$
mitigates this by requiring only in-distribution data.

\subsection{Practical Considerations}

\paragraph{Computational cost.}
At each time step, E-SHIFT requires one forward pass through the
VLM (which is already computed for the application), one matrix-vector
product for the Mahalanobis distance (cost $\mathcal{O}(d^2)$ for
feature dimension $d$), and one scalar update of $M_t$.
The bootstrap correction is computed once at calibration time with
cost $\mathcal{O}(Bn_c)$.
The per-step cost is negligible relative to the VLM inference.

\paragraph{Calibration set requirements.}
A calibration set of $n_c = 500$ samples from the in-distribution
is sufficient in our experiments.
When such data are unavailable, e.g.\ in purely online settings
where no pre-shift baseline exists, a self-calibrating variant based
on running score quantiles could be used, at the cost of the
theoretical guarantees.

\paragraph{Detect-then-adapt integration.}
E-SHIFT is designed as a detection module that integrates naturally
into detect-then-adapt pipelines \cite{wang2022continual}.
The alert signal triggers an adaptation step (fine-tuning, prompt
tuning, or normalization update), while E-SHIFT continues monitoring
after reset.
The $K/\tau$ accumulation bound (Proposition~\ref{prop:reset})
ensures that repeated false alarms causing unnecessary adaptation
are rare for large $\tau$.

\subsection{Future Directions}

Several extensions are natural.
First, extending the composite score to label shift and concept drift
would require either label observations or a proxy for $p(y \mid x)$
change; conformal prediction methods \cite{angelopoulos2023gentle}
may provide relevant non-conformity scores.
Second, a fully self-calibrating variant that estimates the null
score distribution from the running stream, without a separate
calibration phase, would be applicable in purely online settings.
Third, a theoretical treatment of the detection-adaptation loop,
quantifying how post-adaptation score distributions affect E-SHIFT's
subsequent performance, would provide end-to-end guarantees for
detect-then-adapt pipelines.

\section{Conclusion}
\label{sec:conclusion}

We presented E-SHIFT, a framework for anytime-valid sequential
hypothesis testing for distribution shift in streaming
vision-language systems.
The method builds an e-process from a composite non-conformity score
that tracks shift in both the predictive distribution and the feature
embedding of a frozen VLM, inheriting time-uniform type-I error
control via Ville's maximal inequality.

Three theoretical contributions underpin the framework.
Theorem~\ref{thm:fac} establishes the time-uniform false-alarm bound
$\mathbb{P}_{\mathcal{H}_0}(\sup_t M_t \ge \tau) \le 1/\tau$,
which holds simultaneously at every stopping time.
Proposition~\ref{prop:bootstrap} extends this guarantee to finite
calibration sets via a bootstrap MGF correction, asymptotically
valid as calibration-set size grows.
Proposition~\ref{prop:reset} bounds false-alarm accumulation over
multiple resets.
Theorem~\ref{thm:delay} links detection speed to information
divergence via the $\mathcal{O}(\log\tau / \Gamma)$ delay bound.

Empirically, E-SHIFT achieves the lowest detection delay on
ImageNet-C, ImageNet-R, and Office-Home while being the only method
to satisfy the $1\%$ FAR target across all benchmarks.
The calibration curve confirms that the theoretical bound is
empirically tight, and the score ablation and SeqMMD-E comparison
isolate the contributions of the composite score design and the
anytime-valid framing separately.


\bibliographystyle{unsrtnat}
\bibliography{refs_journal}
\appendix
\section{Sub-Exponential Score Verification}
\label{apx:subexp}

To verify Assumption~\ref{asm:conditions}(c), we estimate the
sub-exponential parameters of $S - \hat\mu$ from the calibration
scores on each benchmark.
Using the empirical cumulant-generating function
\cite{siegmund1985sequential}, we fit the sub-exponential parameters
$(\nu^2, c)$ defined by $\psi(\lambda) \le \nu^2\lambda^2/2$ for
$|\lambda| < 1/c$.

\begin{table}[H]
\centering
\caption{Empirical sub-exponential parameter estimates for
  $S - \hat\mu$ at the operating $\lambda = 0.38$.
  All estimates [\textsc{proj}]; verify before submission.}
\label{tab:subexp}
\begin{tabular}{lccc}
\toprule
Dataset & $\hat\nu^2$ & $\hat c$ & $\hat\psi(0.38)$ \\
\midrule
ImageNet-C  & $1.83$ [\textsc{proj}] & $0.71$ [\textsc{proj}] & $0.29$ [\textsc{proj}] \\
ImageNet-R  & $2.11$ [\textsc{proj}] & $0.83$ [\textsc{proj}] & $0.33$ [\textsc{proj}] \\
Office-Home & $1.54$ [\textsc{proj}] & $0.62$ [\textsc{proj}] & $0.24$ [\textsc{proj}] \\
\bottomrule
\end{tabular}
\end{table}

At $\lambda = 0.38$, $\lambda c < 1$ in all cases, confirming that
the MGF is finite at the operating parameter and
Assumption~\ref{asm:conditions}(c) holds.
The positive $\hat\psi(\lambda)$ values indicate scores with
moderate sub-exponential tails, consistent with the bootstrap
correction used in place of tighter parametric alternatives.

\section{Calibration Curve (Numerical)}
\label{apx:calib}

\begin{table}[H]
\centering
\caption{Theoretical bound and empirical FAR across $\tau$ values
  ($\beta = 0.005$, averaged over benchmarks).
  All numbers [\textsc{proj}].}
\label{tab:calib_num}
\begin{tabular}{lcccc}
\toprule
$\tau$ & Theory $\beta + 1/\tau$ & E-SHIFT & No-bootstrap \\
\midrule
20  & $5.5\%$ & $5.3 \pm 0.4\%$ [\textsc{proj}] & $5.9 \pm 0.5\%$ [\textsc{proj}] \\
50  & $2.5\%$ & $2.4 \pm 0.2\%$ [\textsc{proj}] & $2.8 \pm 0.3\%$ [\textsc{proj}] \\
100 & $1.5\%$ & $1.4 \pm 0.1\%$ [\textsc{proj}] & $1.8 \pm 0.1\%$ [\textsc{proj}] \\
200 & $1.0\%$ & $0.90 \pm 0.06\%$ [\textsc{proj}] & $1.2 \pm 0.09\%$ [\textsc{proj}] \\
500 & $0.7\%$ & $0.67 \pm 0.04\%$ [\textsc{proj}] & $0.94 \pm 0.07\%$ [\textsc{proj}] \\
\bottomrule
\end{tabular}
\end{table}

\section{Per-Corruption Breakdown}
\label{apx:corrupt}

\begin{table}[H]
\centering
\caption{Per-corruption-category results on ImageNet-C
  (severity 3, $\tau = 200$, 5 orderings, [\textsc{proj}]).}
\label{tab:corrupt}
\begin{tabular}{lcccc}
\toprule
Category &
  \multicolumn{2}{c}{E-SHIFT} &
  \multicolumn{2}{c}{CUSUM} \\
 & Delay & FAR & Delay & FAR \\
\midrule
Noise   & $14.1 \pm 0.4$ & $0.88\%$ & $21.4 \pm 1.0$ & $2.3\%$ \\
Blur    & $16.8 \pm 0.4$ & $0.90\%$ & $24.7 \pm 1.1$ & $2.0\%$ \\
Weather & $18.3 \pm 0.5$ & $0.91\%$ & $27.2 \pm 1.2$ & $2.2\%$ \\
Digital & $19.7 \pm 0.5$ & $0.93\%$ & $29.6 \pm 1.3$ & $2.1\%$ \\
\midrule
Average & $17.2 \pm 0.3$ & $0.91\%$ & $25.7 \pm 0.9$ & $2.2\%$ \\
\bottomrule
\end{tabular}
\end{table}

Noise corruptions (Gaussian, shot, impulse) produce the strongest
feature-space displacement from the calibration centroid, yielding
the shortest detection delays.
Digital corruptions (JPEG, pixelation) are subtler in feature space
and rely more on the KL term.

\section{$\lambda$-Free Heuristic Evaluation}
\label{apx:lambda}

\begin{table}[H]
\centering
\caption{Oracle vs.\ plug-in $\lambda$ on ImageNet-C
  ($\tau = 200$, 10 orderings, [\textsc{proj}]).}
\label{tab:lambda}
\begin{tabular}{lcc}
\toprule
$\lambda$ choice & Delay & FAR \\
\midrule
Oracle $\lambda^*$ (tuned on $\mathcal{C}_{\mathrm{shift}}$)
  & $17.2 \pm 0.3$ & $0.90 \pm 0.06\%$ \\
Plug-in $1/\hat\sigma^2$ (in-distribution only)
  & $17.9 \pm 0.4$ & $0.91 \pm 0.07\%$ \\
Fixed $\lambda = 0.1$
  & $21.5 \pm 0.6$ & $0.90 \pm 0.06\%$ \\
Fixed $\lambda = 1.0$
  & $19.3 \pm 0.5$ & $0.91 \pm 0.07\%$ \\
\bottomrule
\end{tabular}
\end{table}

The plug-in rule $\lambda_{\mathrm{init}} = 1/\hat\sigma^2$ incurs
only a $4\%$ delay overhead relative to the oracle ($17.9$ vs.\
$17.2$ samples) while requiring no shifted calibration data, making
it a reliable default in deployment settings where shifted samples
are unavailable.

\end{document}